\documentclass[letterpaper, 10 pt, conference]{IEEEconf}  

\IEEEoverridecommandlockouts                              

\usepackage{algorithmic}
\usepackage{rotating}
\usepackage{graphicx}
\usepackage{textcomp}
\usepackage[ruled, algo2e]{algorithm2e}
\usepackage{xcolor}
\usepackage{amsmath,amssymb,amsfonts}
\usepackage{multirow}
\usepackage{soul,color}

\IEEEoverridecommandlockouts
\usepackage{cite}
\def\BibTeX{{\rm B\kern-.05em{\sc i\kern-.025em b}\kern-.08em
    T\kern-.1667em\lower.7ex\hbox{E}\kern-.125emX}}

\title{\LARGE \bf
Enhancing Diversity in Multi-objective Feature Selection\\
\thanks{This research is supported by the Vector Scholarship in Artificial
Intelligence, provided through the Vector Institute. We utilized ChatGPT 3.5 solely for enhancing the writing and performing grammatical checks.}}

 \author{Sevil Zanjani Miyandoab$^{*1}$, Shahryar Rahnamayan$^{*2}$, \emph{SMIEEE}, Azam Asilian Bidgoli$^{*3}$,\\ Sevda Ebrahimi$^{*1}$, Masoud Makrehchi$^{1}$
 \thanks{*Nature-Inspired Computational Intelligence (NICI) Lab}
 \thanks{$^1$Department of Electrical, Computer, and Software Engineering, Ontario Tech University, Oshawa, 
 ON, Canada
          }%
  \thanks{$^2$Department of Engineering, Brock University, St. Catharines, ON, Canada
          }%
  \thanks{$^3$Faculty of Science, Wilfrid Laurier University, Waterloo, ON, Canada
          }%
  \thanks{Contact:{\tt\small sevil.zanjanimiyandoab@ontariotechu.net}
          }%
  }

\usepackage{tikz}
\usepackage{textcomp}
\usepackage{hyperref}
\usepackage{lipsum}

\newcommand\copyrighttext{%
  \footnotesize \textcopyright 2024 IEEE. Personal use of this material is permitted.
  Permission from IEEE must be obtained for all other uses, in any current or future
  media, including reprinting/republishing this material for advertising or promotional
  purposes, creating new collective works, for resale or redistribution to servers or
  lists, or reuse of any copyrighted component of this work in other works.
  DOI: \href{https://ieeexplore.ieee.org/abstract/document/10612084}{10.1109/CEC60901.2024.10612084}
  }
\newcommand\copyrightnotice{%
\begin{tikzpicture}[remember picture,overlay]
\node[anchor=south,yshift=10pt] at (current page.south) {\fbox{\parbox{\dimexpr\textwidth-\fboxsep-\fboxrule\relax}{\copyrighttext}}};
\end{tikzpicture}%
}

\graphicspath{ {./images/} }
\begin{document}

\maketitle
\copyrightnotice
\thispagestyle{empty}
\pagestyle{empty}

\begin{abstract}

Feature selection plays a pivotal role in the data preprocessing and model-building pipeline, significantly enhancing model performance, interpretability, and resource efficiency across diverse domains. In population-based optimization methods, the generation of diverse individuals holds utmost importance for adequately exploring the problem landscape, particularly in highly multi-modal multi-objective optimization problems. Our study reveals that, in line with findings from several prior research papers, commonly employed crossover and mutation operations lack the capability to generate high-quality diverse individuals and tend to become confined to limited areas around various local optima.
This paper introduces an augmentation to the diversity of the population in the well-established multi-objective scheme of the genetic algorithm, NSGA-II. This enhancement is achieved through two key components: the genuine initialization method and the substitution of the worst individuals with new randomly generated individuals as a re-initialization approach in each generation.
The proposed multi-objective feature selection method undergoes testing on twelve real-world classification problems, with the number of features ranging from 2,400 to nearly 50,000. The results demonstrate that replacing the last front of the population with an equivalent number of new random individuals generated using the genuine initialization method and featuring a limited number of features substantially improves the population's quality and, consequently, enhances the performance of the multi-objective algorithm.

\end{abstract}

\section{INTRODUCTION}
Feature selection involves discarding as many features or variables as possible without compromising classification accuracy. The removal of irrelevant and redundant information not only reduces computational requirements but also enhances the classifier's performance by mitigating the curse of dimensionality and facilitates model interpretation~\cite{sevil1, sevil2, chandrashekar2014survey,ghalwash2016structured}.

In recent years, machine learning problems have expanded to include thousands, millions, or even billions of variables, making feature selection a crucial preprocessing step \cite{chandrashekar2014survey}. This practice diminishes data dimensions, simplifying tasks for classifiers and concurrently diminishing memory and computational costs in large-scale datasets. However, excessive feature elimination can lead to classification failure, presenting a delicate balance between the number of features and classification accuracy \cite{sevil1, sevil2, bidgoli2020evolutionary}.

Feature selection finds broad applications across various domains, including machine learning and data science \cite{hall1999correlation, cai2018feature}, bioinformatics and computational biology \cite{saeys2007review, bertolazzi2009application}, text mining \cite{aghdam2009text}, image analysis and medical diagnosis \cite{bidgoli2022evolutionary, asilian2022bias}, natural language processing (NLP) \cite{anand2023deep, suzuki2004convolution}, financial analysis and economics \cite{tsai2009feature, liang2015effect}, and many other fields.

The mainstay of classical feature selection methods falls into three categories: wrapper, filter, and embedded methods \cite{agrawal2021metaheuristic,saeys2007review,guyon2003introduction,chandrashekar2014survey,liu2010feature}. Filter-based methods analyze data characteristics and assess features without engaging learning models \cite{liu2010feature}. Embedded methods incorporate variable selection into the training process, varying for each learning machine. Wrappers score variable subsets based on predictive power under the black-box of the machine \cite{guyon2003introduction,agrawal2021metaheuristic}.

In the realm of multi-objective feature selection algorithms, the Non-dominated Sorting Genetic Algorithm II (NSGA-II) \cite{deb2002fast}, a widely recognized algorithm, has been employed in several research studies \cite{hamdani2007multi, sevil1, sevil2}. This algorithm, a multi-objective scheme of the evolutionary genetic algorithm (GA), comprises three fundamental components: initialization, generating new candidate solutions, and selection. The effectiveness of population-based evolutionary algorithms heavily relies on these components, particularly in providing diverse solutions during optimization. Initialization is a critical step, as confirmed by Kazimipour et al. \cite{kazimipour2014review}, who reviewed popular techniques introduced before 2014 and classified them based on randomness, compositionality, and generality. Rahnamayan et al. \cite{rahnamayan2007novel} proposed an opposition-based initialization method, accelerating optimizer convergence. Burke et al. \cite{burke1998initialization} demonstrated that heuristic initialization methods enhance diversity and performance.

Additionally, the substitution of individuals in the population acts as a form of re-initialization, commonly employed when an evolutionary optimization method converges \cite{ghosh2000genetic}. Ghosh et al. \cite{ghosh2000genetic} highlighted that substituting the worst individuals, and re-entering a dropped-out individual, significantly improves population diversity and, consequently, GA performance. Cobb et al. \cite{cobb1993genetic} introduced Random Immigrants, replacing a fraction of the population randomly in each generation to enhance GA performance. These insights motivated our investigation into the effect of substitution in the multi-objective GA scheme, NSGA-II, applied to feature selection.

In this study, we present a strategy to augment population diversity in multi-objective feature selection algorithms. Our findings underscore a critical point: commonly used crossover and mutation operations lack the potency to generate valuable diverse individuals across the search space. Injecting random individuals from outside the population proves essential for boosting algorithm exploration. In other words, auxiliary re-initialization components are necessary to enhance population diversity. 

The organization of this paper is as follows: Section II provides a detailed background review, Section III outlines our proposed method, illuminating the steps and techniques employed, Section IV presents results and analysis from our study, and Section V concludes with remarks.




\section{Background Review} 
\subsection{Multi-objective Optimization}
Algorithms designed to address multi-objective problems typically navigate trade-off decisions, yielding a set of solutions rather than a singular one. This array of solutions is commonly referred to as the Pareto front or non-dominated solutions. The establishment of the Pareto front is based on the principle of dominance, a criterion used for the comparison of solutions.

\textbf{Definition 1. Multi-objective Optimization}~\cite{asilian2022machine}
\begin{eqnarray}
\begin{aligned}
& Min/Max\  F(\pmb x)=[f_{1}(\pmb x),f_{2}(\pmb x),...,f_{M}(\pmb x)] \\ 
&s.t. \quad L_{i}\leq x_{i}\leq U_{i}, i=1,2,...,d
 \end{aligned}
\end{eqnarray}
Here, $\pmb x$ represents the feature set (i.e., candidate solution), $M$ denotes the number of objectives, $d$ equals the number of variables (i.e., dimension), and each variable's value, denoted as $x_{i}$, falls within the interval $[L_{i}, U_{i}]$. However, in binary multi-objective optimization problems like feature selection, $x_{i}$ is restricted to only two values, either from the set $\{0, 1\}$ or $\{True, False\}$. The objective function, denoted as $f_{i}$, aims for either minimization or maximization~\cite{asilian2022machine}. In such problems, a widely adopted approach for comparing candidate solutions involves the concept of dominance.

\textbf{Definition 2. Dominance Concept} \\
If  $\pmb x=(x_{1},x_{2},...,x_{d})$ and  $ \acute{\pmb x}=(\acute{x}_{1},\acute{x}_{2},...,\acute{ x}_{d})$ are two vectors in a minimization problem search space, $\pmb x$ dominates $\acute{\pmb x}$ ($\pmb x\prec\acute{\pmb x}$) if and only if
\begin{eqnarray}
\begin{aligned}
&\forall i\in{\{1,2,...,M\}}, f_i(\pmb x)\leq f_i(\acute{\pmb x}) \wedge\\ 
&\exists j \in{\{1,2,...,M\}}: f_j(\pmb x)<f_j(\acute{\pmb x})
\end{aligned}
\end{eqnarray}
This concept delineates the optimality of a solution within a multi-objective space. A candidate solution, $\pmb x$, is considered superior to another solution, $\acute{\pmb x}$, if it is not inferior in any of the objectives and, at a minimum, exhibits superior performance in one of the objectives. Solutions that are not dominated by any other form the Pareto front and are termed non-dominated solutions \cite{bidgoli2021reference}.

Another prevalent metric for comparing candidate solutions in multi-objective problems is the crowding distance. In computing the crowding distance, individuals are sorted based on each objective, and boundary individuals (minimums and maximums) receive an infinite distance assignment. Other individuals are assigned the sum of the normalized Euclidean individual distances from their neighbors with the same rank across all objectives. This assesses an individual's importance relative to the density of surrounding individuals~\cite{deb2002fast}.

Multi-objective algorithms strive to identify the Pareto front through the utilization of generating strategies/operators and selection schemes. The non-dominated sorting (NDS) algorithm~\cite{deb2002fast} stands out as a popular selection strategy that operates on the dominance concept. It categorizes the population's solutions into different levels of optimality. The algorithm initiates by identifying all non-dominated solutions in the first rank. To determine the second rank of individuals, non-dominated candidates are removed, and the process repeats for the remaining ones. This iterative procedure continues until all individuals are grouped into distinct Pareto levels. Deb et al. \cite{deb2002fast} introduced the renowned multi-objective optimization method, NSGA-II, based on this concept.

\subsection{Multi-objective Feature Selection}
To transform feature selection into a binary optimization problem, each feature's status is denoted by a binary variable. With the dataset's features having the option of being retained or removed, each variable assumes only two states. Consequently, each individual, or solution, represents a distinct feature configuration. The classification accuracy, calculated by masking the dataset's features and employing a regular classifier such as $k$-Nearest Neighbor ($k$-NN), Support Vector Machine (SVM), or Decision Tree (DT), serves as one objective value for each individual. Additionally, recognizing the significance of the remaining features, especially in large-scale databases, we define the ratio of retained features as the second objective in our optimization problem.

Jiao et al. \cite{jiao2022solving} introduced a multi-objective optimization approach for feature selection in classification, named PRDH. In this work, the authors have devised a duplication handling method to augment diversity in both the objective and search spaces. Their method transforms the multi-objective feature selection problem into a constrained optimization problem, emphasizing classification performance. A novel constraint-handling method is employed to select feature subsets featuring more informative and strongly relevant features.

Cheng et al. \cite{cheng2022variable} demonstrated that the variable granularity search-based multi-objective evolutionary algorithm, VGS-MOEA, significantly narrows down the search space for large-scale feature selection problems. In this method, each bit of the individuals represents a feature subset, with the subset's granularity starting larger and gradually refining, aiming for higher-quality features. This technique markedly enhances feature selection efficiency and accuracy.

In efforts to improve the evaluation and selection of solutions for multi-objective optimization problems, \cite{he2021multiobjective} introduces a novel fitness evaluation mechanism (FEM) utilizing fuzzy relative entropy (FRE). To effectively tackle the problem, a hybrid genetic algorithm is employed within this framework. Deb et al. introduced NSGA-III~\cite{deb2013evolutionary} as a reference-point based many-objective optimization algorithm that prioritizes non-dominated population members in proximity to specified reference points, particularly effective when dealing with four or more objectives. 

More recently, Nikbakhtsarvestani \cite{nikbakhtsarvestani2023multi} proposed a multi-objective coordinate search optimization, outperforming NSGA-II, especially in computationally expensive applications with a limited number of function calls. Zanjani et al. \cite{sevil1} introduced a method for binary multi-objective optimization using coordinate search for feature selection. In this approach, a variable of each individual on the Pareto front is flipped to generate new individuals in each generation. The presented results markedly surpass those of NSGA-II. They also proposed binary Compact NSGA-II \cite{sevil2}, employing Probability Vectors (PVs) to preserve the distribution of solutions, alongside the Pareto front, instead of maintaining two populations during the process.

\section{PROPOSED METHOD}

\subsection{Objectives} 
Enhancing classification accuracy and minimizing data dimensionality constitute the primary objectives of feature selection. Consequently, the calculation of two objectives for each candidate solution (or feature set) becomes imperative. The first objective function computes the classification accuracy, achieved through the utilization of a $k$-NN classifier to assess both the training set (during the optimization phase) and the test set (post-optimization for the evaluation of final solutions).

Given that our optimization problem is of a minimization nature, we calculate the classification error as the first objective, relying on the predictions made by the $k$-NN model:

\begin{equation}
\label{f1}
\mathit{Classification \:Error}= 1 - \frac{\#  Correct\: Predictions}{Total \:\# Predictions}
\end{equation}

The second objective is the diminution of the selected features ratio. Its computation involves tallying the count of features with a value of 1 (or \emph{True}) for each feature set, followed by dividing this count by the total number of features:

\begin{equation}
\label{f2}
\mathit{Ratio\: of\: Selected\: Features}=\frac{Number \:of \: Trues }{Total \:\# Features }
\end{equation}

Both of these objectives possess real values within the interval \([0, 1]\). Prior research has indicated their inherent conflict, so they are suitable candidates for multi-objective optimization, as highlighted in \cite{bidgoli2020evolutionary}.

\subsection{Proposed Algorithm: Binary Diverse NSGA-II}

Despite the effectiveness of NSGA-II, it frequently becomes stuck in local optima and prematurely converges. Our observations suggest that the average pairwise distance among individuals in the feature space is notably low, and as the algorithm progresses, this value declines rapidly. Therefore, a plausible reason for the algorithm's failure to discover the global optimum lies in its insufficient exploration of the search space and the production of diverse solutions.

Based on our experiments and literature findings, the most effective technique for population initialization to enhance diversity and explore the search space is genuine uniform initialization \cite{sevda, bidgoli2020evolutionary}. The commonly used Bit-string Uniform (BU) binary population initialization method fails to provide chromosome-wise uniformity in the population, impacting the quality of the population in the long run. In contrast, the genuine uniform approach or Uniform Covering (UC) binary initialization results in a more diverse population, featuring individuals from various regions of the search space with both chromosome-wise and gene-wise uniformity. In this method, instead of having 50\% \emph{True} variables in nearly all individuals containing a substantial number of variables, the number of \emph{True} variables is uniformly distributed across feature sets.

Subsequently, the objective values of each individual are calculated based on the utilized classifier and Equations \ref{f1}, \ref{f2}. Following this, a non-domination sorting determines the rank of each solution in the population. After generating new individuals - children - through crossover and mutation operations, a non-dominated sorting is executed, and the best individuals survive based on their ranking and crowding distance, akin to the original NSGA-II algorithm.

To address diversity degradation, the proposed idea involves replacing the worst-ranked candidate solutions with newly-generated solutions. At the end of each generation, we replace the last front with new random subsets. These new individuals are generated using the UC initialization method \cite{sevda}, and the number of selected features (i.e., \emph{True} variables) falls within the range of the size of the smallest and largest feature subsets in the current population.

Suppose the size of the smallest (individual with the lowest number of True variables) and largest (individual with the highest number of True variables) feature subsets in the population is given by $\alpha$ and $\beta$, respectively. $[\alpha, \beta]$ determines the range of sizes for reinitialized individuals in the population. For each individual on the last front that must be replaced, a random integer (e.g., $TrueIndices$) within the range of $[\alpha, \beta]$, representing the number of features in the new individual, is generated. Then, $TrueIndices$ number of variables of the new individual are randomly set to \emph{True}, while the others are set to \emph{False}. This constraint, applied to the number of selected features of the new solutions, enhances the algorithm's performance by incorporating prior knowledge from the population and the process, avoiding the generation of entirely random candidate solutions.

Note that in any generation where the entire population is located on a single front, the reinitialization process is bypassed for that generation. 

In contrast to the members generated using crossover and mutation operators, which are in the neighborhood of the previous members and have a small distance from their parents, the new random individuals potentially originate from unseen parts of the landscape and aid in exploring and discovering new areas. A sufficient diversity of members and exploration is particularly critical in vast search spaces.

The termination condition is defined based on the total number of function calls (i.e., evaluations) so that in problems with a demanding evaluation function, such as classification, algorithms can be compared fairly within a specific number of function calls. The number of generations is not an appropriate termination condition because the proposed method's number of evaluations and computations may vary based on the number of individuals on the last front in each generation.

The pseudocode of the proposed binary diverse NSGA-II is presented in Algorithm \ref{alg-one}. Additionally, Algorithm \ref{alg-two} provides details on the genuine initialization method or UC.

\begin{algorithm2e}
\SetAlgoLined
\SetKwInOut{Input}{input}\SetKwInOut{Output}{output}
 \Input{ $dataset$, $N$, $NVars$, $maxNFC$ }
 \tcp{$N$: population size}
 \tcp{$NVars$: number of variables}
 \Output{ $Pareto Front ~Solutions$ }
 \BlankLine

$population = $ GenuineInitialization ($N$, $NVars$, 1, $NVars$)\;
Evaluate ($population$)\;

\While{Number of Evaluations $<$ $maxNFC$}{
    $NDS~(population)$\;
    $parents$ = selection ($population$)\;
    $children$ = crossover ($parents$)\;
    $children$ = mutation ($children$)\;
    Evaluate ($children$)\;
    $population$ = Merge ($population$, $children$)\;
    $NDS~(population)$\;
    Calculate Crowding Distance ($population$)\;
    $population$ = Survive $N$ best individuals\;
    $N' = len~ (LastFront)$\;
    \If {$N' < N$}
    {
    
    $NumTruesArr$ = an array containing the number of \emph{True} variables for each individual\;
    $a$ = the minimum value in $NumTruesArr$\;
    $b$ = the maximum value in $NumTruesArr$\;
    $NewSolutions = $ GenuineInitialization ($N'$, $NVars$, $a$, $b$) \;
    Evaluate ($NewSolutions$)\;
    Replace ($LastFront$, $NewSolutions$)\;}
}
\BlankLine
\caption{Binary diverse NSGA-II}\label{alg-one}
\end{algorithm2e}

\begin{algorithm2e}
\SetAlgoLined
\SetKwInOut{Input}{input}\SetKwInOut{Output}{output}
 \Input{ $N$, $NVars$ , $minVars$, $maxVars$}
 \tcp{$N$: population size}
 \tcp{$NVars$: number of variables}
 \Output{ $NewIndividuals$ }
\BlankLine
$NewIndividuals$ = 2d array size [$N$, $NVars$] filled with $False$\;
\For{$i\leftarrow 1$ \KwTo $N$}{
    $NumberOfTrues$ = a random integer in range $[minVars, maxVars]$\;
    $TrueIndices$ = random $NumberOfTrues$ samples from range $[1, NVar]$ without replacement\;
    $NewIndividuals$ [$i$, $TrueIndices$] = $True$\;
}

\BlankLine
\caption{Genuine Initialization}\label{alg-two}
\end{algorithm2e}

\section{EXPERIMENTAL RESULTS AND ANALYSIS}
\subsection{Datasets}
We have evaluated the efficacy of our proposed approach (i.e., diverse NSGA-II), NSGA-II, and the genuinely initialized version of NSGA-II, across twelve extensive datasets within the domains of microarray and image analysis. The objective is to identify the optimal feature set. These datasets are characterized by a substantial number of features paired with a relatively limited number of instances. Consequently, the reduction of features is imperative to enhance classification accuracy. The properties of the employed datasets are outlined in Table~\ref{tab-datasets}.

\begin{table}[htbp]
\caption{Datasets Description}
\begin{center}
\begin{tabular}{|c|c|c|c|c|}
\hline
\textbf{Number} & \textbf{Dataset} & \textbf{\#features} & \textbf{\#samples} & \textbf{\#classes} \\ \hline
D1              & warpAR10P \cite{zhao2010advancing}        & 2400                & 130                  & 10                 \\ \hline
D2              & warpPIE10P \cite{zhao2010advancing}       & 2420                & 210                  & 10                 \\ \hline
D3              & LUNG \cite{bhattacharjee2001classification}             & 3312                & 203                  & 5                  \\ \hline
D4              & GLIOMA \cite{nutt2003gene}           & 4434                & 50                   & 4                  \\ \hline
D5              & TOX-171 \cite{zhao2010advancing}          & 5748                & 171                  & 4                  \\ \hline
D6              & LEUKEMIA \cite{golub1999molecular}         & 7070                & 72                   & 2                  \\ \hline
D7              & CARCINOM \cite{su2001molecular}        & 9182                & 174                  & 11                 \\ \hline
D8              & pixraw10P \cite{zhao2010advancing}        & 10000               & 100                  & 10                 \\ \hline
D9              & CLL-SUB-111 \cite{zhao2010advancing}      & 11340               & 111                  & 3                  \\ \hline
D10             & SMK-CAN-187 \cite{zhao2010advancing}      & 19993               & 187                  & 2                  \\ \hline
D11             & GLI-85 \cite{zhao2010advancing}           & 22283               & 85                   & 2                  \\ \hline
D12             & GLA-BRA-180 \cite{zhao2010advancing}      & 49151               & 180                  & 4                  \\ \hline
\end{tabular}
\label{tab-datasets}
\end{center}
\end{table}

\subsection{Experimental Settings}
We conducted multiple runs of the algorithms, repeating the process 31 times with different seeds to mitigate the impact of stochasticity. In each iteration, a distinct twenty percent of instances from each dataset were randomly chosen as a test set. So, the method of validation is Repeated Random Subsampling validation or Monte Carlo Cross-Validation. Recognizing the computational expense associated with feature selection on large-scale datasets, we maintained a fixed number of function calls at 15,000 for the mentioned algorithms, ensuring a fair comparison. Our experiments primarily rely on the pymoo framework \cite{pymoo}. Further details on hyperparameter settings can be found in Table~\ref{tab-settings}.

\begin{table}[htbp]
\caption{Parameter Settings for both proposed method and NSGA-II algorithms}
\begin{center}
\begin{tabular}{|ll|}
\hline
\multicolumn{1}{|l|}{Population size}                & 100                    \\ \hline
\multicolumn{1}{|l|}{Number of function calls (NFC)} & 15,000                  \\ \hline
\multicolumn{1}{|l|}{Selection method}               & Tournament Selection   \\ \hline
\multicolumn{1}{|l|}{Mutation method}                & Bit-flip Mutation      \\ \hline
\multicolumn{1}{|l|}{Mutation probability}           & 0.01                   \\ \hline
\multicolumn{1}{|l|}{Crossover}                      & Single Point Crossover \\ \hline
\multicolumn{1}{|l|}{Survival method}                & NDS and crowding distance\\ \hline
\multicolumn{1}{|l|}{Duplicate Elimination}          & TRUE                   \\ \hline
\multicolumn{1}{|l|}{Number of runs}                 & 31                     \\ \hline
\end{tabular}
\label{tab-settings}
\end{center}
\end{table}

Our objective is to minimize both the classification error and the ratio of selected features to all features, both of which fall within the real value interval [0,1]. A suitable multi-objective evaluation metric for comparing algorithm performance is hypervolume (HV), with the reference point set at $(1, 1)$. Furthermore, to calculate the classification error, we utilize the $k$-NN classifier with a fixed parameter $k$ set to 5 across all datasets and experiments.

\subsection{Numerical Results and Analysis}

\begin{figure*}
\centering
\begin{tabular}{ccc}
\includegraphics[width=0.30\linewidth]{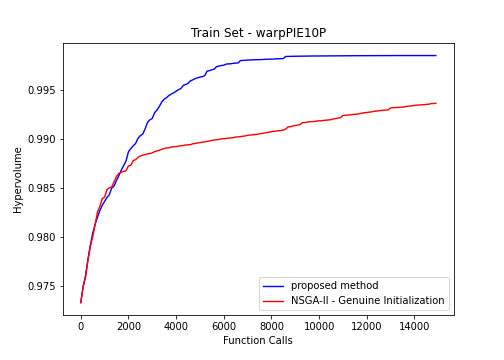}
&\includegraphics[width=0.30\linewidth]{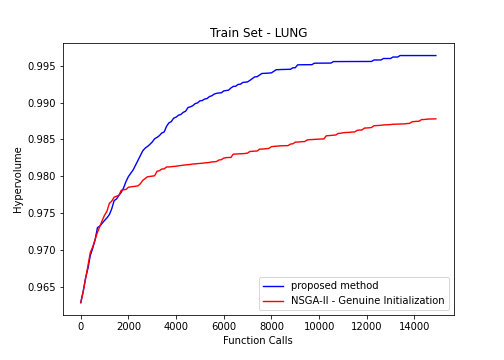}
&\includegraphics[width=0.30\linewidth]{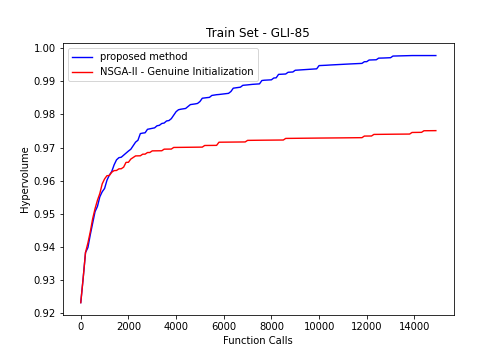}
 \\
(a) warpPIE10P & (b) LUNG & (c) GLI-85 \\


\end{tabular}
\caption{Comparing HV plots during optimization for the traditional and proposed diverse NSGA-II on some datasets}
\label{img-1}
\end{figure*}

Table \ref{tab-results-HV} shows that across all datasets, our method notably surpasses regular NSGA-II in terms of the HV performance metric. Average HV values during optimization for NSGA-II with genuine initialization (red) and the proposed binary diverse NSGA-II method (blue) for sample datasets are depicted in Fig.~\ref{img-1}. The genuine initialization establishes a markedly superior starting point for the HV value compared to the regular NSGA-II, and the substitution of low-ranked individuals accelerates HV growth by promoting exploration of the search space. The combination of these techniques facilitates a rapid HV increase, resulting in an average value of 0.97, as shown in Table \ref{tab-results-HV}. Notably, for half of the datasets, the mean train HV exceeds 0.99, with none falling below 0.89. Conversely, NSGA-II fails to attain a train HV of 0.67 or higher for any dataset.

Furthermore, we observed that, in most cases, the diverse NSGA-II generates a greater number of candidate solutions on the Pareto front, all featuring fewer features compared to solutions on NSGA-II's Pareto front. Table \ref{tab-results-HV}, which includes the average final HV values for the train and test Pareto fronts after 15,000 function calls, confirms the statistically significant superiority of the proposed method's numerical results against NSGA-II across all cases. These results have been validated using a statistical t-test at a 95\% confidence level. Notably, the HV value on the test set consistently exceeds 0.70 for all datasets when employing our proposed method, while the maximum test HV value for the regular NSGA-II is 0.58. Moreover, in 9 out of 12 cases, the test HV values obtained from our proposed method surpasses that of the genuinely initialized NSGA-II, and we see three statistical ties.

This table also underscores the substantial impact of the initialization approach on the final HV value, surpassing the effect of solution substitution. Specifically, employing the genuine initialization enhances the train set HV by approximately 0.42 and the test set HV by about 0.34. The addition of a straightforward method to diversify the population further, such as the replacement technique, yields an average improvement of approximately 0.04 on the test set HV.

\begin{table*}[htbp]
\caption{Final train HV values of NSGA-II, NSGA-II with Genuine Initialization, and
diverse NSGA-II. \# w/t/l stands for number of wins/ties/losses based on the t-test statistical test with a confidence level of 95\%.}
\begin{center}
\begin{tabular}{|c|ccc|ccc|}
\hline
\textbf{}        & \multicolumn{3}{c|}{\textbf{Train HV}}                                                                                                                                         & \multicolumn{3}{c|}{\textbf{Test HV}}                                                                                                                                           \\ \hline
\textbf{Dataset} & \multicolumn{1}{c|}{\textbf{NSGA-II}} & \multicolumn{1}{c|}{\textbf{\begin{tabular}[c]{@{}c@{}}NSGA-II with\\ Genuine Initialization\end{tabular}}} & \textbf{\textbf{\begin{tabular}[c]{@{}c@{}}Proposed\\ Diverse NSGA-II\end{tabular}}} & \multicolumn{1}{c|}{\textbf{NSGA-II}} & \multicolumn{1}{c|}{\textbf{\begin{tabular}[c]{@{}c@{}}NSGA-II with \\ Genuine Initialization\end{tabular}}} & \textbf{\textbf{\begin{tabular}[c]{@{}c@{}}Proposed\\ Diverse NSGA-II\end{tabular}}} \\ \hline
D1               & \multicolumn{1}{c|}{0.5294}           & \multicolumn{1}{c|}{\textbf{0.9572}}                                                                                 & \textbf{0.9643}          & \multicolumn{1}{c|}{0.3470}           & \multicolumn{1}{c|}{0.6472}                                                                                  & \textbf{0.7052}          \\ \hline
D2               & \multicolumn{1}{c|}{0.6692}           & \multicolumn{1}{c|}{0.9937}                                                                                 & \textbf{0.9985}          & \multicolumn{1}{c|}{0.4566}           & \multicolumn{1}{c|}{0.7354}                                                                                  & \textbf{0.7674}          \\ \hline
D3               & \multicolumn{1}{c|}{0.6267}           & \multicolumn{1}{c|}{0.9878}                                                                                 & \textbf{0.9964}          & \multicolumn{1}{c|}{0.5833}           & \multicolumn{1}{c|}{\textbf{0.9017}}                                                                                  & \textbf{0.9044}          \\ \hline
D4               & \multicolumn{1}{c|}{0.5420}           & \multicolumn{1}{c|}{0.9689}                                                                                 & \textbf{0.9965}          & \multicolumn{1}{c|}{0.4176}           & \multicolumn{1}{c|}{0.7058}                                                                                  & \textbf{0.7546}          \\ \hline
D5               & \multicolumn{1}{c|}{0.5543}           & \multicolumn{1}{c|}{0.9829}                                                                                 & \textbf{0.9889}          & \multicolumn{1}{c|}{0.4175}           & \multicolumn{1}{c|}{\textbf{0.7673}}                                                                                  & \textbf{0.7927}          \\ \hline
D6               & \multicolumn{1}{c|}{0.5630}           & \multicolumn{1}{c|}{0.9916}                                                                                 & \textbf{0.9998}          & \multicolumn{1}{c|}{0.4942}           & \multicolumn{1}{c|}{0.8416}                                                                                  & \textbf{0.9439}          \\ \hline
D7               & \multicolumn{1}{c|}{0.5159}           & \multicolumn{1}{c|}{0.9581}                                                                                 & \textbf{0.9748}          & \multicolumn{1}{c|}{0.4176}           & \multicolumn{1}{c|}{0.7475}                                                                                  & \textbf{0.7815}          \\ \hline
D8               & \multicolumn{1}{c|}{0.5517}           & \multicolumn{1}{c|}{0.9914}                                                                                 & \textbf{0.9999}          & \multicolumn{1}{c|}{0.3973}           & \multicolumn{1}{c|}{\textbf{0.7239}}                                                                         & \textbf{0.7080}                   \\ \hline
D9               & \multicolumn{1}{c|}{0.4663}           & \multicolumn{1}{c|}{0.9414}                                                                                 & \textbf{0.9603}          & \multicolumn{1}{c|}{0.3814}           & \multicolumn{1}{c|}{0.7524}                                                                                  & \textbf{0.8151}          \\ \hline
D10              & \multicolumn{1}{c|}{0.4430}           & \multicolumn{1}{c|}{0.8863}                                                                                 & \textbf{0.9183}          & \multicolumn{1}{c|}{0.4178}           & \multicolumn{1}{c|}{0.7974}                                                                                  & \textbf{0.8328}          \\ \hline
D11              & \multicolumn{1}{c|}{0.5066}           & \multicolumn{1}{c|}{0.9751}                                                                                 & \textbf{0.9978}          & \multicolumn{1}{c|}{0.4586}           & \multicolumn{1}{c|}{0.8634}                                                                                  & \textbf{0.9045}          \\ \hline
D12              & \multicolumn{1}{c|}{0.4122}           & \multicolumn{1}{c|}{0.8572}                                                                                 & \textbf{0.8957}          & \multicolumn{1}{c|}{0.3876}           & \multicolumn{1}{c|}{0.7534}                                                                                  & \textbf{0.7959}          \\ \hline
\textbf{Average} & \multicolumn{1}{c|}{0.5317}           & \multicolumn{1}{c|}{0.9576}                                                                                 & \textbf{0.9743}          & \multicolumn{1}{c|}{0.4314}           & \multicolumn{1}{c|}{0.7698}                                                                                  & \textbf{0.8088}          \\ \hline
\textbf{\# w/t/l} & \multicolumn{1}{c|}{0/0/12}                & \multicolumn{1}{c|}{0/1/11}                                                                                      & \textbf{-}              & \multicolumn{1}{c|}{0/0/12}                & \multicolumn{1}{c|}{0/3/9}                                                                                       & \textbf{-}              \\ \hline
\end{tabular}
\label{tab-results-HV}
\end{center}
\end{table*}

\begin{figure*}
\centering
\begin{tabular}{ccc}
\includegraphics[width=0.30\linewidth]{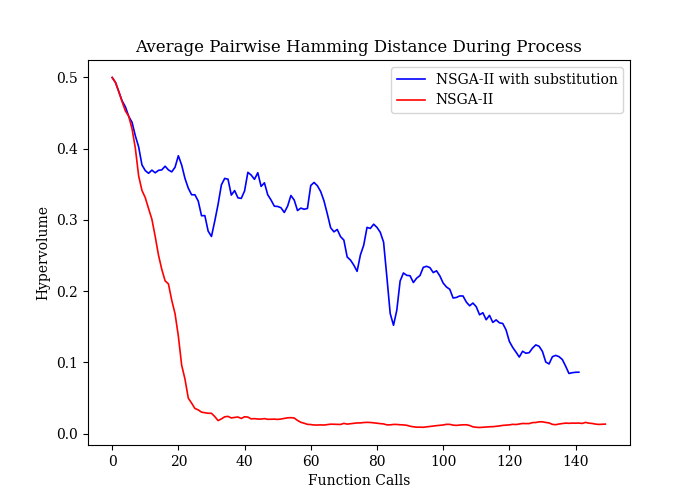}
&\includegraphics[width=0.30\linewidth]{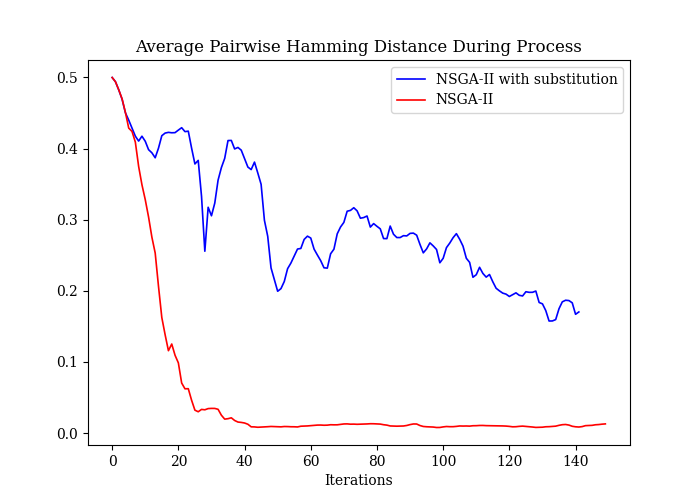}
&\includegraphics[width=0.30\linewidth]{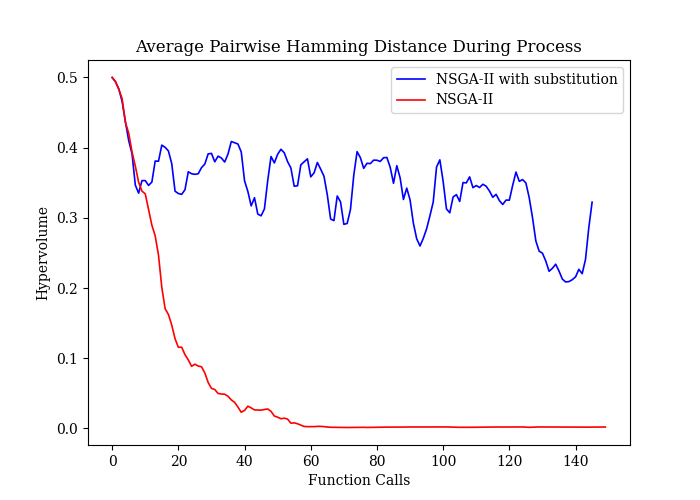}
 \\
(a) warpPIE10P & (b) LUNG & (c) GLI-85 \\


\end{tabular}
\caption{Average pairwise Hamming distance plots during optimization for
NSGA-II with and without the substitution component}
\label{img-div}
\end{figure*}

Fig. \ref{img-div} illustrates the average pairwise Hamming distance plots obtained during optimization using NSGA-II with and without the substitution component. The plots highlight that substituting low-ranked individuals significantly increases the distance between individuals during the process, thereby enhancing diversity. Both methods employed random BU initialization for a fair comparison in this experiment.

Table \ref{tab-max-acc} presents the average of the maximum accuracy over the 31 runs and the ratio of the features selected in the most accurate solutions. These results have also been validated using a statistical t-test at 95\% confidence level. Across all datasets, except for LUNG and pixraw10P (D3 and D8), our method significantly enhances the classifier's accuracy compared to the regular NSGA-II. For the mentioned two datasets, we see statistical ties. In essence, looking at it from a singular objective perspective, diverse NSGA-II method consistently achieves superior accuracy as well. This heightened accuracy is accomplished using less than 1\% of the total features, a substantial reduction compared to the number of features selected by NSGA-II. The average number of selected features in the most accurate solution found by NSGA-II across various datasets is approximately 5508, whereas the improved scheme identifies better solutions with about 164 times less memory requirement, averaging 33.5 variables. Consequently, we benefit from both improved accuracy and reduced memory consumption, potentially expediting pipelined processes.

Furthermore, Table \ref{tab-max-acc} illustrates that the impact of substitution in improving algorithm performance to find a more accurate solution is almost comparable to the initialization method. On average, it elevates maximum accuracy by over 3\% compared to NSGA-II with genuine initialization, concurrently reducing the number of features by over three times.

\begin{table*}[htbp]
\caption{Average of the maximum accuracy of the individuals and the ratio of features on the most accurate solutions. \# w/t/l stands for number of wins/ties/losses based on the t-test statistical test with a confidence level of 95\%.}
\begin{center}
\begin{tabular}{|c|ccc|ccc|}
\hline
\textbf{}        & \multicolumn{3}{c|}{\textbf{Maximum Accuracy}}                                                                                                                                  & \multicolumn{3}{c|}{\textbf{\#Features in the Most Accurate Solution}}                                                                           \\ \hline
\textbf{Dataset} & \multicolumn{1}{c|}{\textbf{NSGA-II}} & \multicolumn{1}{c|}{\textbf{\begin{tabular}[c]{@{}c@{}}NSGA-II with \\ Genuine Initialization\end{tabular}}} & \textbf{\textbf{\begin{tabular}[c]{@{}c@{}}Proposed\\ Diverse NSGA-II\end{tabular}}} & \multicolumn{1}{c|}{\textbf{NSGA-II}} & \multicolumn{1}{c|}{\textbf{\begin{tabular}[c]{@{}c@{}}NSGA-II with\\ Genuine Initialization\end{tabular}}} & \textbf{\textbf{\begin{tabular}[c]{@{}c@{}}Proposed\\ Diverse NSGA-II\end{tabular}}} \\ \hline
D1               & \multicolumn{1}{c|}{0.5347}           & \multicolumn{1}{c|}{0.6514}                                                                                  & \textbf{0.7060}          & \multicolumn{1}{c|}{846.2258}         & \multicolumn{1}{c|}{22.8065}                                                                                & \textbf{10.1290}         \\ \hline
D2               & \multicolumn{1}{c|}{0.6759}           & \multicolumn{1}{c|}{0.7389}                                                                                  & \textbf{0.7680}          & \multicolumn{1}{c|}{792.1935}         & \multicolumn{1}{c|}{15.1290}                                                                                & \textbf{7.9032}          \\ \hline
D3               & \multicolumn{1}{c|}{\textbf{0.9111}}  & \multicolumn{1}{c|}{\textbf{0.9072}}                                                                                  & \textbf{0.9048}                   & \multicolumn{1}{c|}{1191.5161}        & \multicolumn{1}{c|}{22.5484}                                                                                & \textbf{6.5806}          \\ \hline
D4               & \multicolumn{1}{c|}{0.6774}           & \multicolumn{1}{c|}{0.7097}                                                                                  & \textbf{0.7548}          & \multicolumn{1}{c|}{1700.1935}        & \multicolumn{1}{c|}{25.4516}                                                                                & \textbf{2.3548}          \\ \hline
D5               & \multicolumn{1}{c|}{0.7152}           & \multicolumn{1}{c|}{\textbf{0.7733}}                                                                                  & \textbf{0.7935}          & \multicolumn{1}{c|}{2400.0323}        & \multicolumn{1}{c|}{59.8387}                                                                                & \textbf{25.9355}         \\ \hline
D6               & \multicolumn{1}{c|}{0.8473}           & \multicolumn{1}{c|}{0.8473}                                                                                  & \textbf{0.9441}          & \multicolumn{1}{c|}{2951.6452}        & \multicolumn{1}{c|}{47.6452}                                                                                & \textbf{1.8387}          \\ \hline
D7               & \multicolumn{1}{c|}{0.7429}           & \multicolumn{1}{c|}{0.7539}                                                                                  & \textbf{0.7825}          & \multicolumn{1}{c|}{4032.3226}        & \multicolumn{1}{c|}{99.9677}                                                                                & \textbf{39.3548}         \\ \hline
D8               & \multicolumn{1}{c|}{\textbf{0.6952}}           & \multicolumn{1}{c|}{\textbf{0.7290}}                                                                         & \textbf{0.7081}                   & \multicolumn{1}{c|}{4284.9355}        & \multicolumn{1}{c|}{75.6452}                                                                                & \textbf{1.7097}          \\ \hline
D9               & \multicolumn{1}{c|}{0.6886}           & \multicolumn{1}{c|}{0.7588}                                                                                  & \textbf{0.8163}          & \multicolumn{1}{c|}{5075.7419}        & \multicolumn{1}{c|}{111.7419}                                                                               & \textbf{28.3548}         \\ \hline
D10              & \multicolumn{1}{c|}{0.7725}           & \multicolumn{1}{c|}{0.8048}                                                                                  & \textbf{0.8345}          & \multicolumn{1}{c|}{9190.0323}        & \multicolumn{1}{c|}{192.9355}                                                                               & \textbf{57.4194}         \\ \hline
D11              & \multicolumn{1}{c|}{0.8482}           & \multicolumn{1}{c|}{0.8710}                                                                                  & \textbf{0.9051}          & \multicolumn{1}{c|}{10241.7419}       & \multicolumn{1}{c|}{199.7742}                                                                               & \textbf{20.2903}         \\ \hline
D12              & \multicolumn{1}{c|}{0.7392}           & \multicolumn{1}{c|}{0.7608}                                                                                  & \textbf{0.7984}          & \multicolumn{1}{c|}{23394.9355}       & \multicolumn{1}{c|}{513.2258}                                                                               & \textbf{200.4516}        \\ \hline
\textbf{Average} & \multicolumn{1}{c|}{0.7374}           & \multicolumn{1}{c|}{0.7755}                                                                                  & \textbf{0.8097}          & \multicolumn{1}{c|}{5508.4600}        & \multicolumn{1}{c|}{115.5591}                                                                               & \textbf{33.5269}         \\ \hline
\textbf{\# w/t/l} & \multicolumn{1}{c|}{0/2/10}                & \multicolumn{1}{c|}{0/3/9}                                                                                       & -              & \multicolumn{1}{c|}{0/0/12}                & \multicolumn{1}{c|}{0/0/12}                                                                                                   & -             \\ \hline
\end{tabular}
\label{tab-max-acc}
\end{center}
\end{table*}

Table \ref{CI-obj} displays the 95\% confidence intervals for the number of selected features for the solutions on the Pareto front. According to these results, the average value of the interval for the second objective obtained from the proposed method is significantly superior to that of the other models. This finding further confirms the effectiveness of both employed techniques in enhancing feature reduction.

\begin{table}[htbp]
\caption{95\% Confidence Interval of the objective values on the Pareto front solutions of each algorithm. \# w/t/l stands for number of wins/ties/losses. }
\setlength{\tabcolsep}{2.5pt}
\begin{center}
\begin{tabular}{|c|ccc|}
\hline
\textbf{}        & \multicolumn{3}{c|}{\textbf{\#Features}}                                                                                                                                       \\ \hline
\textbf{Dataset} & \multicolumn{1}{c|}{\textbf{NSGA-II}} & \multicolumn{1}{c|}{\textbf{\begin{tabular}[c]{@{}c@{}}NSGA-II with\\ Genuine Initialization\end{tabular}}} & \textbf{\textbf{\begin{tabular}[c]{@{}c@{}}Proposed\\ Diverse NSGA-II\end{tabular}}} \\ \hline
D1               & \multicolumn{1}{c|}{(834, 851)}       & \multicolumn{1}{c|}{(13.17, 19.17)}                                                                         & \textbf{(4.10, 6.28)}    \\ \hline
D2               & \multicolumn{1}{c|}{(779, 811)}       & \multicolumn{1}{c|}{(8.19, 11.33)}                                                                          & \textbf{(3.51, 4.69)}    \\ \hline
D3               & \multicolumn{1}{c|}{(1181, 1198)}     & \multicolumn{1}{c|}{(13.69, 23.17)}                                                                         & \textbf{(3.17, 4.51)}    \\ \hline
D4               & \multicolumn{1}{c|}{(1687, 1703)}     & \multicolumn{1}{c|}{(21.66, 45.25)}                                                                         & \textbf{(1.56, 2.18)}    \\ \hline
D5               & \multicolumn{1}{c|}{(2380, 2406)}     & \multicolumn{1}{c|}{(31.15, 46.06)}                                                                         & \textbf{(9.37, 16.11)}   \\ \hline
D6               & \multicolumn{1}{c|}{(2939, 2969)}     & \multicolumn{1}{c|}{(35.63, 62.43)}                                                                         & \textbf{(1.40, 1.81)}    \\ \hline
D7               & \multicolumn{1}{c|}{(4014, 4049)}     & \multicolumn{1}{c|}{(50.99, 77.58)}                                                                         & \textbf{(13.13, 21.49)}  \\ \hline
D8               & \multicolumn{1}{c|}{(4276, 4299)}     & \multicolumn{1}{c|}{(58.18, 100.40)}                                                                        & \textbf{(1.36, 1.69)}    \\ \hline
D9               & \multicolumn{1}{c|}{(5068, 5115)}     & \multicolumn{1}{c|}{(68.23, 115.15)}                                                                        & \textbf{(10.72, 32.41)}  \\ \hline
D10              & \multicolumn{1}{c|}{(9166, 9207)}     & \multicolumn{1}{c|}{(155.30, 246.32)}                                                                       & \textbf{(23.41, 59.59)}  \\ \hline
D11              & \multicolumn{1}{c|}{(10222, 10261)}   & \multicolumn{1}{c|}{(121.84, 228.92)}                                                                       & \textbf{(6.52, 22.72)}   \\ \hline
D12              & \multicolumn{1}{c|}{(23392, 23450)}   & \multicolumn{1}{c|}{(286.43, 490.89)}                                                                       & \textbf{(92.23, 197.59)} \\ \hline
\textbf{Average} & \multicolumn{1}{c|}{(5495, 5526)}     & \multicolumn{1}{c|}{(72.04, 122.22)}                                                                        & \textbf{(14.21, 30.92)}  \\ \hline
\textbf{\# w/t/l} & \multicolumn{1}{c|}{0/0/12}                & \multicolumn{1}{c|}{0/0/12}                                                                                      & \textbf{-}              \\ \hline
\end{tabular}
\label{CI-obj}
\end{center}
\end{table}

\begin{figure*}
\centering
\begin{tabular}{ccc}
\includegraphics[width=0.30\linewidth]{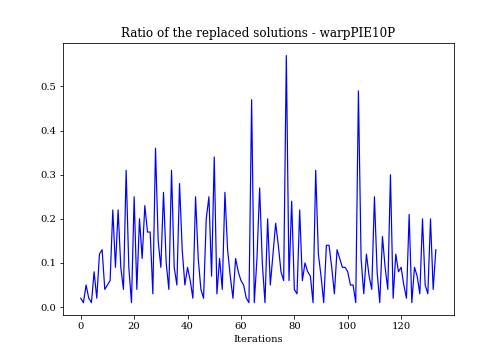}
&\includegraphics[width=0.30\linewidth]{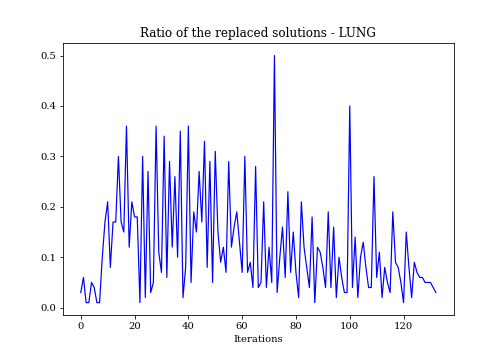}
&\includegraphics[width=0.30\linewidth]{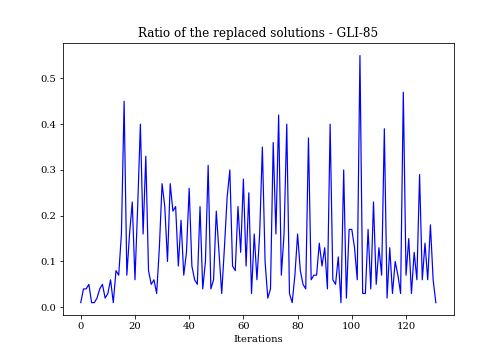}
 \\
(a) warpPIE10P & (b) LUNG & (c) GLI-85\\


\end{tabular}
\caption{Sample plots of the ratio of the replaced individuals in each generation of diverse NSGA-II over generations.}
\label{len-inits}
\end{figure*}

The average ratio of replaced solutions over the generations in the proposed method is also presented in Table \ref{tab-rep}. On average, 11\% of the solutions are positioned on the last front and replaced in each generation. Remarkably, this percentage remains almost consistent across all twelve datasets, showing no significant variations. Fig. \ref{len-inits} includes several sample plots depicting the ratio of replaced solutions over the generations. As shown in the plots, the number of solutions on the last front exhibits no discernible correlation with the generation number or the dataset overall, with the curve fluctuating throughout the generations. However, the initial points of the curves are generally lower, indicating higher diversity in the initial population. The curve occasionally drops to zero when all solutions are concentrated on a single front, resulting in no solution replacements in that generation.

\begin{table}[htbp]
\caption{Average Ratio of the Replaced Solutions}
\begin{center}
\begin{tabular}{|c|c|}
\hline
\textbf{Dataset} & \textbf{Replaced Solutions \%} \\ \hline
D1               & 10.98                                         \\ \hline
D2               & 11.80                                         \\ \hline
D3               & 11.80                                         \\ \hline
D4               & 11.02                                         \\ \hline
D5               & 11.58                                         \\ \hline
D6               & 11.58                                         \\ \hline
D7               & 11.81                                         \\ \hline
D8               & 11.00                                         \\ \hline
D9               & 10.88                                         \\ \hline
D10              & 10.09                                         \\ \hline
D11              & 11.21                                         \\ \hline
D12              & 10.06                                         \\ \hline
\textbf{Average} & \textbf{11.15}                                \\ \hline
\end{tabular}
\label{tab-rep}
\end{center}
\end{table}


\section{CONCLUSION REMARKS}

With the growing importance of feature selection in large-scale real-world datasets, there's an increasing demand for efficient binary multi-objective optimization algorithms. These algorithms should not only minimize memory and computational demands but also reduce classification errors or enhance other processing tasks.
In this research, we improved the population diversity of the NSGA-II algorithm through a novel approach, resulting in enhanced performance. Our method replaced the Bit-string Uniform binary population initialization with a more effective genuine initialization strategy, significantly increasing population diversity. Additionally, we proposed substituting candidate solutions on the last front in each generation to further diversify the population and efficiently explore the search space. Our findings demonstrated that these techniques not only substantially reduce the number of selected features but also maintain or boost classification accuracy, thereby increasing the Hypervolume (HV) of both train and test sets. The complexity of this method matches that of NSGA-II. One drawback of replacing solutions is that we may lose some information and potentially beneficial solutions.

Our proposed method is tailored for binary multi-objective optimization in general, with feature selection serving as a specific case study. However, its utility extends far beyond feature selection to address a wide range of binary optimization challenges. Moreover, the demonstrated benefits of dimension reduction observed in this study, particularly in enhancing classification and $k$-NN performance, can also be harnessed to improve efficiency and effectiveness across various other processing tasks. Furthermore, given that crossover and mutation operations in genetic algorithms like NSGA-II do not generate new individuals from diverse parts of the landscape, we recommend exploring alternative techniques to enhance population diversity as a promising avenue for future research.

\bibliography{ref}
\bibliographystyle{IEEEtran}

\addtolength{\textheight}{-12cm}   

\end{document}